\pdfoutput=1

\documentclass[11pt]{article}

\usepackage{acl}

\usepackage{times}
\usepackage{latexsym}
\usepackage{amsmath}

\usepackage{algorithm}
\usepackage{algorithmic}
\usepackage[T1]{fontenc}

\usepackage[utf8]{inputenc}

\usepackage{microtype}

\usepackage{inconsolata}
\usepackage{multirow}
\usepackage{graphicx}
\usepackage{subcaption}    
\title{Fast Chain-of-Thought: A Glimpse of Future From Jacobi Decoding Leads to Answers Faster}

\author{
Hongxuan Zhang \\ Nanjing University\\ 
  \And
  Zhining Liu \\ Ant Group \\
  \And
  Yao Zhao \\ Ant Group\\
 \And
  Jiaqi Zheng \\ Nanjing University \\  
  \AND
  Chenyi Zhuang  \\ Ant Group \\
  \And
  Jinjie Gu\\ Ant Group \\ 
  \And
  Guihai Chen\\ Nanjing University \\
  \AND
  \texttt{x\_zhang@smail.nju.edu.cn}
  }

\begin{document}
\maketitle
\begin{abstract}

CoT (chain of thought) is widely used in reasoning tasks with large language models (LLM). In this work, we propose FastCoT, a model-agnostic framework to accelerate the inference of CoT tasks without training or modification. FastCoT utilizes parallel decoding and autoregressive decoding simultaneously. Parallel decoding generates multiple approximate tokens in a forward, and autoregressive decoding leverages these preliminary approximate tokens to yield one or more refined tokens. Distinct from conventional approaches that rely solely on precisely generated tokens from autoregressive decoding, FastCoT integrates these approximate yet informative tokens in the final response generation process. The approximate tokens act as \textbf{a quick glance} of the future tokens, which could lead to faster generation compared to regular autoregressive decoding. Through extensive experiments, we demonstrate that FastCoT accelerates inference by nearly $20\%$ on wide models, with only a negligible performance drop compared to the regular approach.


\end{abstract}

\section{Introduction}

The NLP field has undergone a revolution with the introduction of large language models (LLMs) consisting of tens or hundreds of billions of parameters. These models are pretrained on large-scale corpora using self-supervised training objectives. One notable milestone in this domain is the release of ChatGPT, a powerful AI chatbot that is developed based on LLMs and has garnered widespread attention from society. Following this breakthrough, several other large-scale language models have been introduced, including 
Llama \cite{touvron2023llama1,touvron2023llama}, PaLM \cite{chowdhery2022palm}, and Bloom \cite{Scao2022BLOOMA1},  all of which have achieved remarkable success in areas such as language understanding and text generation.

\begin{figure}[h]
    \centering
    \includegraphics[width=1\linewidth]{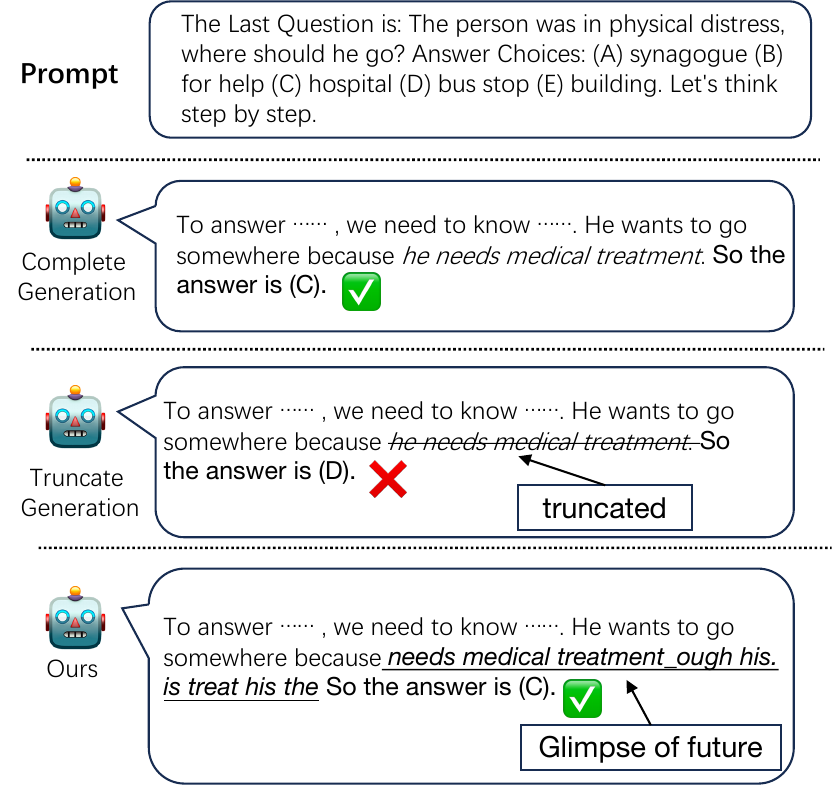}
    \caption{Example of how the glimpse of future work in CoT reasoning task. In the CoT reasoning task, LLM is required to completely generate the complete rationale and finally get the answer based on the rationale. FastCoT argues that partial generation of the complete rationale with a glimpse of future is enough. The example in this figure comes from our experiment results.}
    \label{fig: introduction of motivation}
\end{figure}

However, several challenging issues still remain unresolved in the field, including arithmetic, commonsense, and symbolic reasoning. In order to enhance the performance of LLMs on various tasks, the Chain of Thoughts (CoT) series of work has been proposed. The concept of CoT prompting was first introduced by \cite{wei2022chain}. This series of work aims to improve the performance of LLMs on a wide range of reasoning tasks by guiding the model to generate its own rationales, which are a series of intermediate reasoning steps. Notice that this performance improvement comes at the cost of generating additional rationales. Since the main focus is often on obtaining the answers, the extra generation of rationales may be seen as a drawback. Furthermore, most state-of-the-art causal transformers are autoregressive models, meaning they can only predict tokens one at a time. This leads to slow inference and may not fully utilize the capabilities of GPUs\cite{Yang2023InferenceWR}.


The nature of autoregressive models imposes significant computational load, which can be limiting for certain real-world applications, especially for industrial applications that involve large amounts of data flow and require high throughput, which would lead to high costs and poor user experience. Therefore, a natural question arises: can a large language model (LLM) benefit from its inner approximate reasoning rationale? It is noteworthy that humans usually think a lot in mind before getting started to write down on paper. Although these vague ideas or thoughts may not be fully connected to form complete sentences, they can still be beneficial for the reasoning task. As shown in Figure \ref{fig: introduction of motivation}, we argue that word pieces generated by a fast but inaccurate decoding strategy are informative to generate correct results. 


To validate our arguments, we experimented by corrupting the rationale generated by the LLM, and led the LLM to infer the answer according to the corrupted rationale instead. For more details please refer to Section \ref{sec: corrupting rationale}. A major observation is that even a randomly sampled oracle rationale can lead to correct answers, indicating that the LLMs can make accurate predictions without having access to the complete rationale. Motivated by this, we propose a method called FastCoT, which can utilize approximate rationale to find answers faster. This method combines the exact tokens obtained through any lossless decoding methods with the \textit{approximate} tokens obtained through Jacobi decoding. By doing so, we can obtain the answer to the reasoning task with fewer forward inferences of the LLM. Our contributions can be summarized in three main aspects:



\begin{enumerate}
    \item We first introduce Jacobi decoding into the reasoning task, such as CoT, and propose to use the by-products of Jacobi decoding, which are approximate tokens, as a glimpse of the future for LLM in the decoding reasoning process. We demonstrate and analyze how Jacobi decoding reduces time overhead in reasoning task scenarios from two aspects: The first aspect comes from Jacobi decoding itself, which can reduce the number of iterations by generating more than one token in a single forward inference with probability. The second aspect comes from a glimpse into the future through approximate tokens, which can help LLM analyze the final answer in advance without autoregressively decoding it.

    \item We conduct extensive experiments on LLMs with different scales and datasets, showing speedups of up to $20\%$ in inference time. In the reasoning task scenario, we analyze the time overhead of each part compared to the most commonly used lossless method, i.e., autoregressive decoding. To the best of our knowledge, this is the first study to accelerate inference of CoT tasks with inaccurate reasoning steps.

    \item Third, we have designed a parallel decoding framework to support a wide range of large language models with almost no modification to the source code of the language model itself based on the huggingface implementations.
\end{enumerate}

\section{Related Work}

\noindent\textbf{XoT Prompt Engineering} The Chain-of-Thought (CoT) prompting \cite{wei2022chain,kojima2022large} is proposed to induce large-scale language models to think step-by-step, similar to how humans approach complex questions. A typical few-shot CoT prompt consists of K examples with corresponding rationales and answers for the questions. After that, several methods have been proposed to enhance the performance of CoT prompting across various domains and tasks, including Self-Ask \cite{press2022measuring}, Self-Consistency \cite{wang2022self}, Self-Verification \cite{Weng2022LargeLM}, Maieutic Prompting \cite{Jung2022MaieuticPL}, Automate-CoT \cite{Shum2023AutomaticPA}, MoT \cite{Li2023MoTME}, ToT \cite{long2023large}, and GoT \cite{besta2023graph}. Additionally, \cite{Zhang2022AutomaticCO} proposes a clustering-based approach to automatically select questions from different classes to form a CoT Prompt. \cite{Diao2023ActivePW} propose using a metric called uncertainty to identify the most uncertain problems in LLM and enhance CoT's performance by providing artificial answers to those questions. These tasks are referred to as XoT. Our work is parallel to these engineering efforts and can be applied to any prompting method related to XoT. In terms of utilizing rationale in XoT Prompting, several works, such as \cite{Magister2022TeachingSL, Hsieh2023DistillingSO, Li2023SymbolicCD, Wang2023SCOTTSC}, attempt to use CoT-generated rationale to improve the task performance of student models across diverse datasets through rationale distillation. 
These methods are capable of yielding more accurate results compared to the non-XoT approach. However, they compromise the speed of inference. To the best of our knowledge, we are the first to propose exploiting the concept of approximate rationale to expedite the completion of XoT-series tasks, and our method is orthogonal to the prompt engineering approach.

\noindent \textbf{Accuracy Lossless Decoding Acceleration}


Many works have proposed various methods for accelerating the generation of autoregressive causal language models. \cite{stern2018blockwise} introduced Blockwise Parallel Decoding, utilizing K language-model-dependent decoding heads to predict K more tokens compared to autoregressive decoding. On the other hand, \cite{kasai2020deep} proposed using a deeper Encoder and fewer layers of Decoder to achieve faster speed. \cite{xia2022speculative, leviathan2023fast, chen2023accelerating} suggested employing a faster but less powerful draft model to assist large models in generating multiple tokens in one transformer call. In the field of Machine Translation, \cite{gu2017non, huang2022non} introduced Non-Autoregressive Translation (NAT) models to overcome the limitations of autoregressive decoding. However, the application of NAT models is mostly limited to machine translation and related domains, and needs careful parameter tuning.



The most closely related work to ours is \cite{santilli2023accelerating}, where they propose the use of Jacobi decoding in translation tasks. However, they do not leverage any approximate tokens generated during Jacobi decoding, and they do not provide an implementation that supports batch computation. To the best of our knowledge, we are the first paper to attempt to utilize these approximate tokens. Compared to the methods mentioned earlier in this section, our framework does not require any additional training on either the base pre-trained language model or a new model.



%
%

\section{Preliminary}

\subsection{Parallel Jacobi Decoding}
For a trained model with parameters $\theta$, a forward inference of the LLM would generate the distribution $p_{\theta}(y_i|y_{1:i-1}, x)$ for the next token based on the prompt $x$ and the tokens generated so far $y_{1:i-1}$. In the case of a common greedy autoregressive decoding approach without considering any post logits-processor, the next token would be selected by choosing the token with the highest probability from the distribution, which can be represented as below.


\begin{equation}\label{eqn:autoregressive}
\begin{aligned}
    y_{i} \leftarrow & \mathop{\arg\max}p_{\theta}(y_i | y_{1:i-1} , x) 
\end{aligned}
\end{equation}

In contrast to autoregressive decoding, Jacobi decoding \cite{santilli2023accelerating} takes into account additional tokens after the generating position, resulting in a corresponding increase in the number of outputs. To facilitate the description, we refer to the additional tokens considered during forward inference in this context as the \textit{context window}, denoted its size as $c$. A Jacobi decoding process can be shown as,

\begin{equation}
    \begin{cases}
        y_{i}   &\leftarrow \mathop{\arg\max}p_{\theta}(y_i | x, y_{1:i-1})   \\
        \widehat{Y}^{t}_{i+1} &\leftarrow \mathop{\arg\max}p_{\theta}(y_{i+1} |x, y_{1:i-1}, \widehat{Y}^{t-1}_{i})     \\ 
        \widehat{Y}^{t}_{i+2} &\leftarrow \mathop{\arg\max}p_{\theta}(y_{i+2} |x, y_{1:i-1}, \widehat{Y}^{t-1}_{i:i+1})     \\ 
        
        & \vdots    \\
        \widehat{Y}^{t}_{i+c} &\leftarrow \mathop{\arg\max}p_{\theta}(y_{i+c} | x, y_{1:i-1}, \widehat{Y}^{t-1}_{i:i+c-1})
     \end{cases}
\end{equation}

\noindent For the simplicity of subsequent formulation, we denote one iteration of the Jacobi Decoding as,

\begin{equation}
    y_{i}, \widehat{Y}^{t} \leftarrow JD(x, y_{<i}, \widehat{Y}^{t-1}, i, c )
\end{equation}

\noindent where $i$ is used to indicate the start position of the Jacobi decoding, and we use $y$ and $\widehat{Y}$ to distinguish the tokens corresponding to the accuracy-lossless decoding part and the Jacobi decoding part. Variable $t$ is used to differentiate the results produced by each iteration. The $y^{t-1} $ means the iterative solution of the previous iteration, which is also the $t^{th}$ iteration's input to the LLM. We name $\widehat{Y}^{t}$ as approximate tokens due to they cannot pass the verification and therefore cannot be accepted as text generated by lossless decoding, they are considered as \textbf{by-products} of exact parts in previous work. Compared with the previous description of the Jacobi decoding from \cite{santilli2023accelerating}, we specially add the superscript $t$ to highlight the role of approximate tokens generated during the iterative process in our method. 




\begin{figure}
    \centering
    \includegraphics[width=\linewidth]{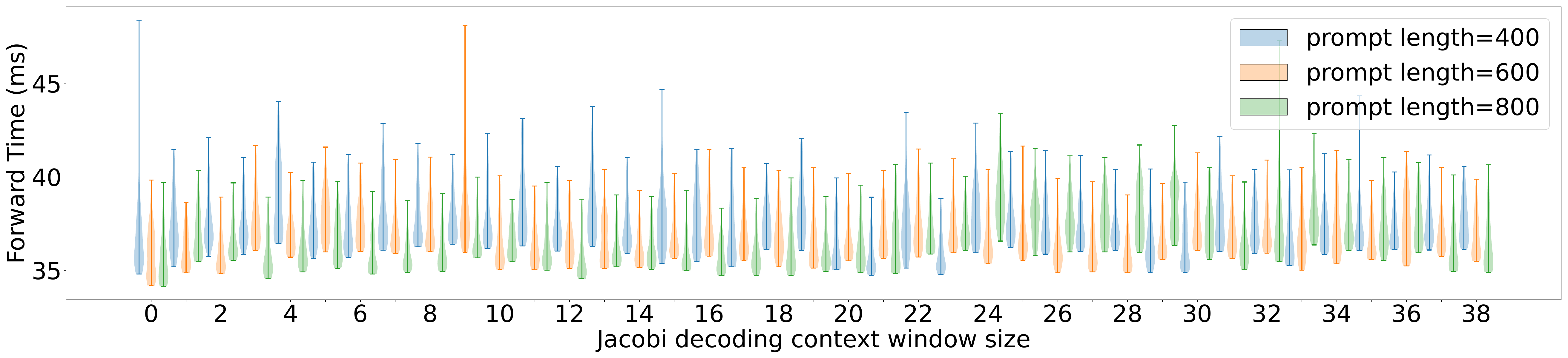}
    \caption{Time cost with context window size $c$. The time overhead does not increase significantly as the window size increases. Almost no change in time consumption from autoregressive decoding (size=0) to Jacobi decoding with a window size of 38. We apply the Llama2-13B model with a single Nvidia A100 GPU.}
    \label{fig:time cost of different c}
\end{figure}

\noindent\textbf{Is context window time-consuming?} Intuitively, increasing the number of tokens that the model needs calculate during inference will lead to an increase in latency. However, we will demonstrate counter-intuitive results through experiments: that is, the context window within a certain limit will not cause an increase in inference latency. We conducted a single inference task to measure the time overhead for a single forward inference under different window sizes, while considering different prompt lengths. Figure \ref{fig:time cost of different c} illustrates the results, the forward time almost remains constant with respect to the context length. The primary contributory factor underlying this phenomenon is that, when the decoding length is comparatively limited, the consequent diminutive matrix blocks can not fully leverage the computational capabilities of CUDA and Tensor cores to their fullest extent. Our findings suggest that increasing the window size within a certain range has only a minimal impact on latency for Jacobi decoding. 
%



\section{Method}


\begin{figure}[h]
    \centering
    \includegraphics[width=1\linewidth]{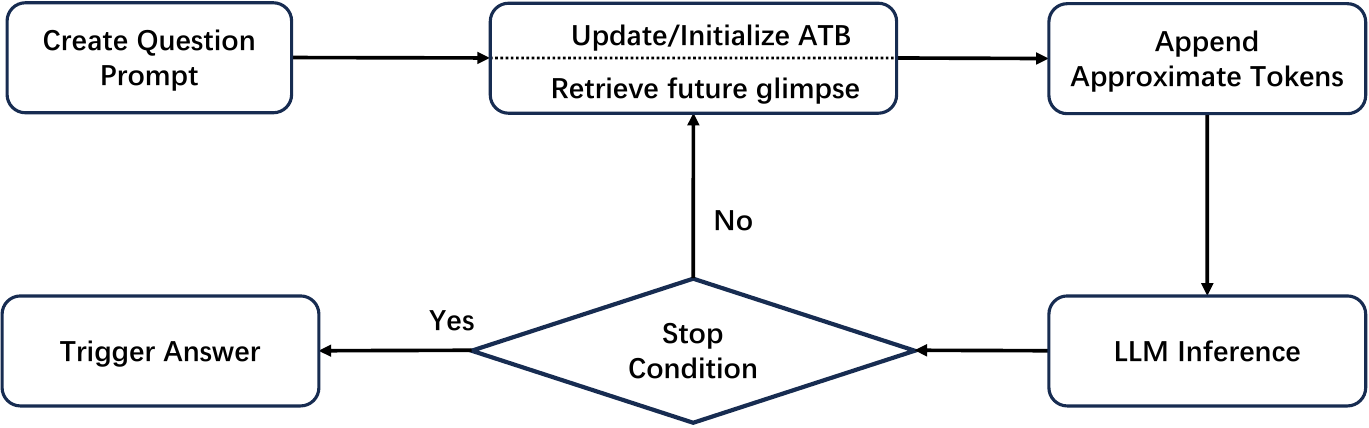}
    \caption{Overview of FastCoT. The autoregressive decode get another one exact token, further future’s approximate token is revealed by Jacobi decoding.}
    \label{fig:overview}
\end{figure}

In our research, we interrogate whether Jacobi decoding by-products can offer a \textbf{glimpse of the future}. Current applications and methods in LLMs, such as those employing Jacobi decoding for machine translation (\cite{santilli2023accelerating}), focus exclusively on precise decoding outcomes, neglecting the potential insights from by-products as they cannot be verified during forward propagation. We thought about whether the by-products part of Jacobi Decoding could help with certain tasks. Considering that by-products mostly consist of tokens related to the final generated text, we call it a glimpse of the future sentence. We would like to emphasize the importance of this \textit{glimpse} into the future, 
which could play an important role in enabling faster answer generation in XoT-series scenarios.

To begin with, we will briefly introduce the important components of our proposed algorithm. Then, we will introduce how to iterate through approximate tokens through Jacobi decoding.

\subsection{Overview of FastCoT}

First, in a manner similar to the original CoT methodology, FastCoT would construct a prompt for the question $Q$ that requires reasoning by LLM. Secondly, FastCoT sequentially retrieves approximate tokens from the \textbf{Approximate Tokens Buffer} to fill the context window. Then these tokens are appended to the tokenized prompt, thereby establishing an initial input for Jacobi decoding processes. Third, the LLM would conduct the inference call. After this step, at least one token precisely matches that which would be produced by autoregressive decoding, thus facilitating the generation of a solution for the subsequent iteration. The newly derived approximate tokens would be used to update the corresponding entries within the Approximate Tokens Buffer. Finally, upon meeting the termination criterion, all generated tokens (including those approximated) are amalgamated with the answer trigger to elicit the final response from the LLM.


\subsection{Approximate Tokens Buffer}
In contrast to the original CoT rationales elicited by LLMs in standard settings, the rationales employed in FastCoT consist of two distinct components: a precise segment that is identical to the output generated via any accuracy-lossless decoding, and an ambiguous segment comprising \textit{approximate tokens} produced through Jacobi decoding. These approximate tokens, derived from Jacobi decoding, may not precisely match those procured through lossless decoding processes, because of the lack of verification from language models. In our approach, we use $Y$ to refer the Approximate Tokens Buffer, where $Y=[ Y_{0:I}^{t}, \widehat{Y}^{t}_{I:} ]$. Here $Y_{0:I}^{t}$ means the exact tokens and $\widehat{Y}^{t}_{I:}$ represents the approximate part. The vector $I$ contains a batch-size number of integers indicating the positions of the first approximate tokens in each instance of the batch. As the decoding process iterates, the approximate tokens in the buffer are gradually transformed into exact tokens. When $t=0$, FastCoT initialize $I$ as $\mathbf{0}$. We choose to initialize $Y$ with populating tokenized token id sequence of the question itself based on empirical results for each question. Compared with the commonly used initialization method, i.e., initializing using a special token [PAD], our method would provide a more diverse initial solution for early stage's Jacobi decoding.

\subsection{Approximate Rationale Generation}
\label{sec:vague rationale generation}

\begin{figure}
    \centering
    \includegraphics[width=1\linewidth]{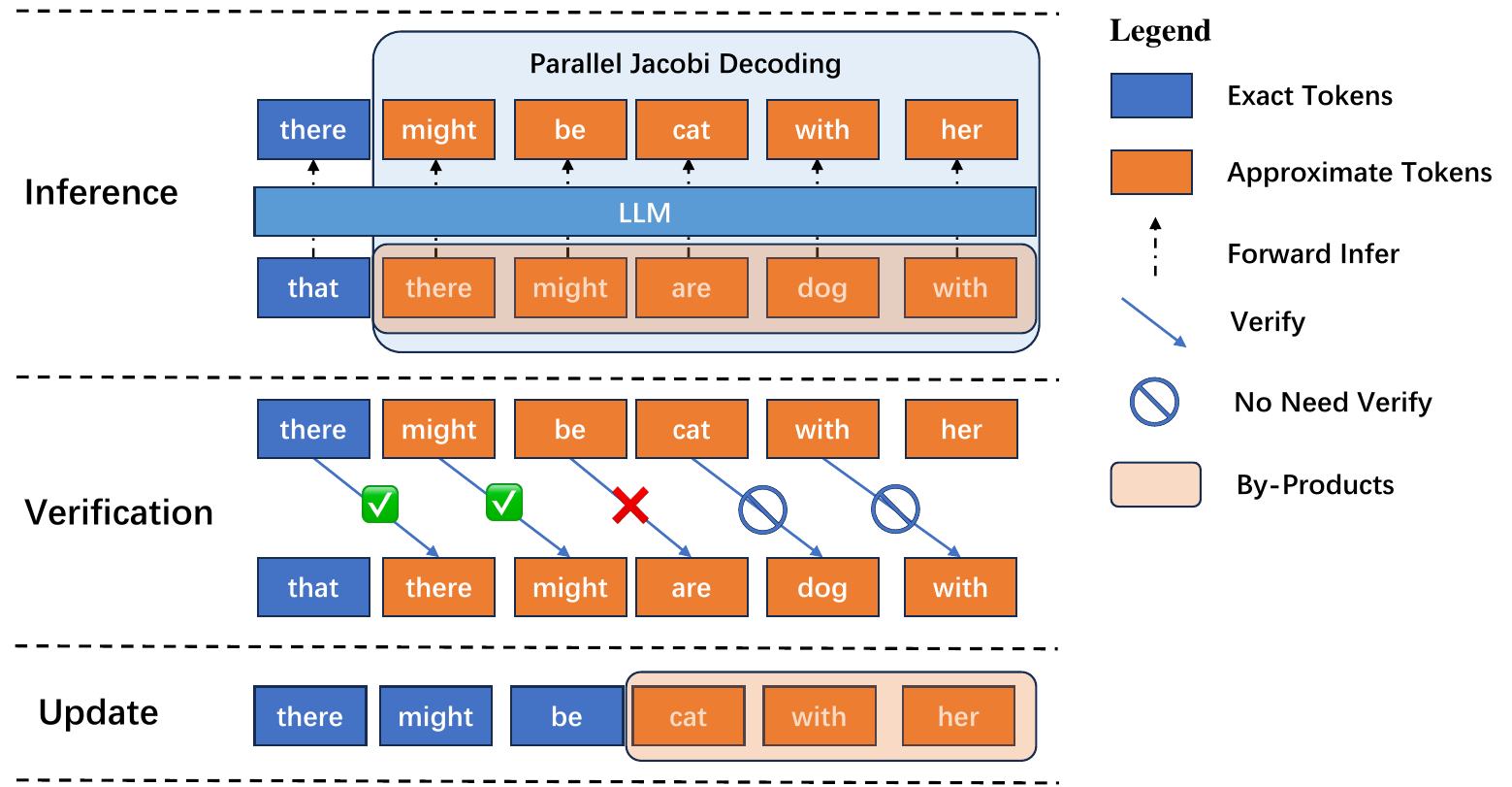}
    \caption{Overview of a complete iteration of FastCoT.}
    \label{fig:FastCoT iteration}
\end{figure}

We will divide the explanation into three parts as shown in Figure \ref{fig:FastCoT iteration}.

\noindent\textbf{FastCoT Inference}:
During the iterative inference process, lossless decoding and Jacobi decoding are performed in parallel within a single forward, leveraging the causal attention mask used in causal language models. The expression is as follows:

\begin{equation}\label{eqn:inference}
    y_{I}, \widehat{Y}_{I+1:I+c+1}^{t+1} \leftarrow PD(x, \textbf{y}_{<I}, \widehat{Y}^{t}_{I:I+c}, I, c )
\end{equation}

\noindent \textbf{Verification}:
In contrast to autoregressive decoding, where only one exact token is generated during each inference step, Jacobi decoding has the ability to generate more simultaneously. So we have to determine the index of the last exact token within the approximate tokens buffer by figuring out,
\begin{equation}\label{eqn:verification}
\begin{aligned}
    K \leftarrow \mathop{\arg\max}\limits_{K}( Y^{t+1}_{I:I+K} & = Y^{t}_{I:I+K})
\end{aligned}
\end{equation}
\noindent as shown in the Figure \ref{fig:FastCoT iteration}'s Verification part, validate one by one until the incorrect one is identified. This index $I$ represents the position up to which the tokens can be considered as exact tokens, while any tokens beyond that index are approximate (Although they can also be correct after further verification).

\noindent \textbf{Cache Update}: After the verification process is completed, we update all the results obtained in this iteration's forward inference to the Approximate Tokens Buffer. The tokens located within $I^{t}+K$ would be accepted as the verified exact token solution, while the excluded tokens must remain as approximate tokens. These approximate tokens will enter the next iteration's forward inference as iterative solutions or serve as the vague rationale for the final CoT answer. After that, $I^{t}$ would be updated to $I^{t+1}+K$, where $K$ is calculated by Equation \ref{eqn:verification}.



\subsection{Iteration Stop Condition}
Because the performance of the CoT task will gradually converge with the iterative process of Jacobi decoding, we have designed different iteration termination conditions to determine when to terminate under various circumstances. Once any of these conditions are met, the iteration will be terminated. 
\begin{enumerate}
    \item For a large dataset, we randomly select a small portion of it as $\mathcal{S}_{cal}$ and perform the CoT task within this subset for each iteration. We calculate the minimum number of iterations required to achieve a certain performance loss threshold and use this value as the upper bound for iterations.\label{condition: upper bound}
    \item EOS (End Of Sentence): EOS has been verified or decoded by accuracy-lossless decoding indicating the end of text generation.
\end{enumerate}

\noindent \textbf{Trigger Answer} Relying solely on vague rationales may not be sufficient to prompt the LLM to generate answers in advance due to LLM would generate tokens continually. To address this, we have designed a specific prompt called answer trigger. The answer trigger serves as a prompt that explicitly instructs the LLM to output the answer to the question. This trigger remains consistent if this part exists in the prompt, such as in few-shot setting prompts. When the iteration termination condition is triggered, we combine the complete prompt, along with all generated tokens, and include the answer trigger, to instruct the LLM to generate the final answer directly.

\section{Experiment}

\subsection{Experiments Settings}

\noindent \textbf{Datasets and Language Models}
We use three widely used datasets for reasoning tasks, including CSQA~\cite{talmor2018commonsenseqa}, StrategyQA~\cite{geva2021did}, AQuA~\cite{ling2017program}. During evaluations, the official test split for each dataset is used. 
And we conduct experiments on the following models: Llama2-13B, Llama2-7B, Llama-13B, Llama-7B~\cite{touvron2023llama}.

\noindent \textbf{Evaluation}
Performance evaluation is bifurcated into two key metrics: accuracy and efficiency. Accuracy is gauged by the proportion of correct responses, while efficiency is discerned through the metrics of wall-clock time and iteration count. Our inference environment includes a server with a 32-core CPU, 64 GiB host memory, and an A100-SXM(80G) GPU. We implemented our framework based on the transformers library of Huggingface\footnote{https://huggingface.co/}

For CoT prompt, we remain consistent with the settings in the active prompt paper. Regarding sliding window size control, we keep the context window length fixed throughout the iteration. In our experiments, we use the following sentence as answer trigger, \textit{"So the answer is"}. 
In the experiment, we will answer the following questions:
\begin{itemize}
    \item Q1: Whether the approximate rationales help the reasoning task get the correct answer? 
    \item Q2: Whether FastCoT can achieve a fast inference speed?
    \item Q3: What is the difference between FastCoT decoding time composition and autoregressive decoding?
    \item Q4: How context window influence the downstream CoT task during the iteration process?
\end{itemize}

\noindent \textbf{Comparison Methods} To answer the above questions, we have designed the following baseline and experiments: 

\begin{itemize}
    \item Vanilla CoT, using the autoregressive decoding method(The abbreviation is AR). 
    \item FastCoT(w/o by-products), autoregressive decoding with truncate generated rationale to iteration number. The generation of CoT answers would based on the original prompt and the truncated rationales. Compared with FastCoT, the only thing FastCoT(w/o by-products) does not have is the approximate part tokens, so we regard it as a very strong baseline.
    \item FastCoT, differs from FastCoT (w/o by-products) in that, given an equal number of iterations, FastCoT maintains the same count of exact tokens but adds an additional number of approximate tokens proportional to the window size.
\end{itemize}

\subsection{Corrupting Rationale Experiment}
\label{sec: corrupting rationale}

We design an experiment utilizing partially corrupted rationales, to verify our argument that in many reasoning tasks in various fields, complete rationales are not necessary, and having some keywords or key information instead can be sufficient to guide the LLM in obtaining the correct answer.

In this experiment, we initially conducted the Vanilla CoT task to elicit the corresponding rationale for each question. To simulate partial information scenarios, we introduced a predefined ratio of [PAD] tokens to substitute original tokens within the rationale. We employed two distinct random patterns for this substitution, each occurring with equal likelihood. The first pattern involved sequentially masking tokens from the rationale's end towards the beginning, a method intended to emulate partial generation as depicted in Figure \ref{fig: introduction of motivation}. The second pattern randomly and uniformly obscured tokens across the rationale, designed to mimic the outcomes of Jacobi decoding. We iterated this masking process using 100 distinct random seeds to create a variety of corrupted rationales for the identical question. Subsequently, we fused the original prompt with each corrupted rationale and presented it as input to the same LLM to derive an answer informed by the corrupted rationale.

\begin{figure}[t]
    \centering
    \includegraphics[width=\linewidth]{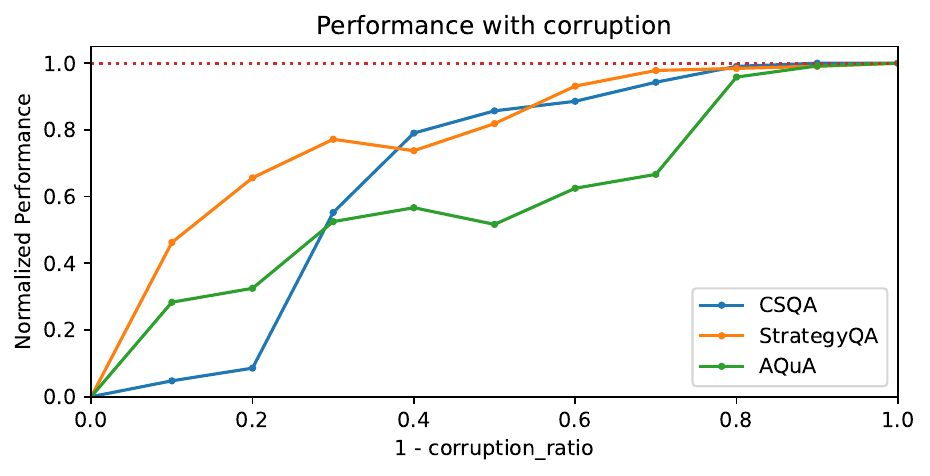}
    \caption{ Reasonable overlap ratio would lead to a saturated accuracy. }
    \label{fig:Exp corrupting rationale}
\end{figure}

As shown in Figure \ref{fig:Exp corrupting rationale}, a reasonable overlap ratio would lead to a saturated accuracy. For instance, when only $40\%$ of the rationale generated by autoregressive decoding is revealed in StrategyQA dataset, the performance is already saturated.

\subsection{Performance Analysis}

\begin{figure}[t]
    \centering
    \includegraphics[width=0.98\linewidth]{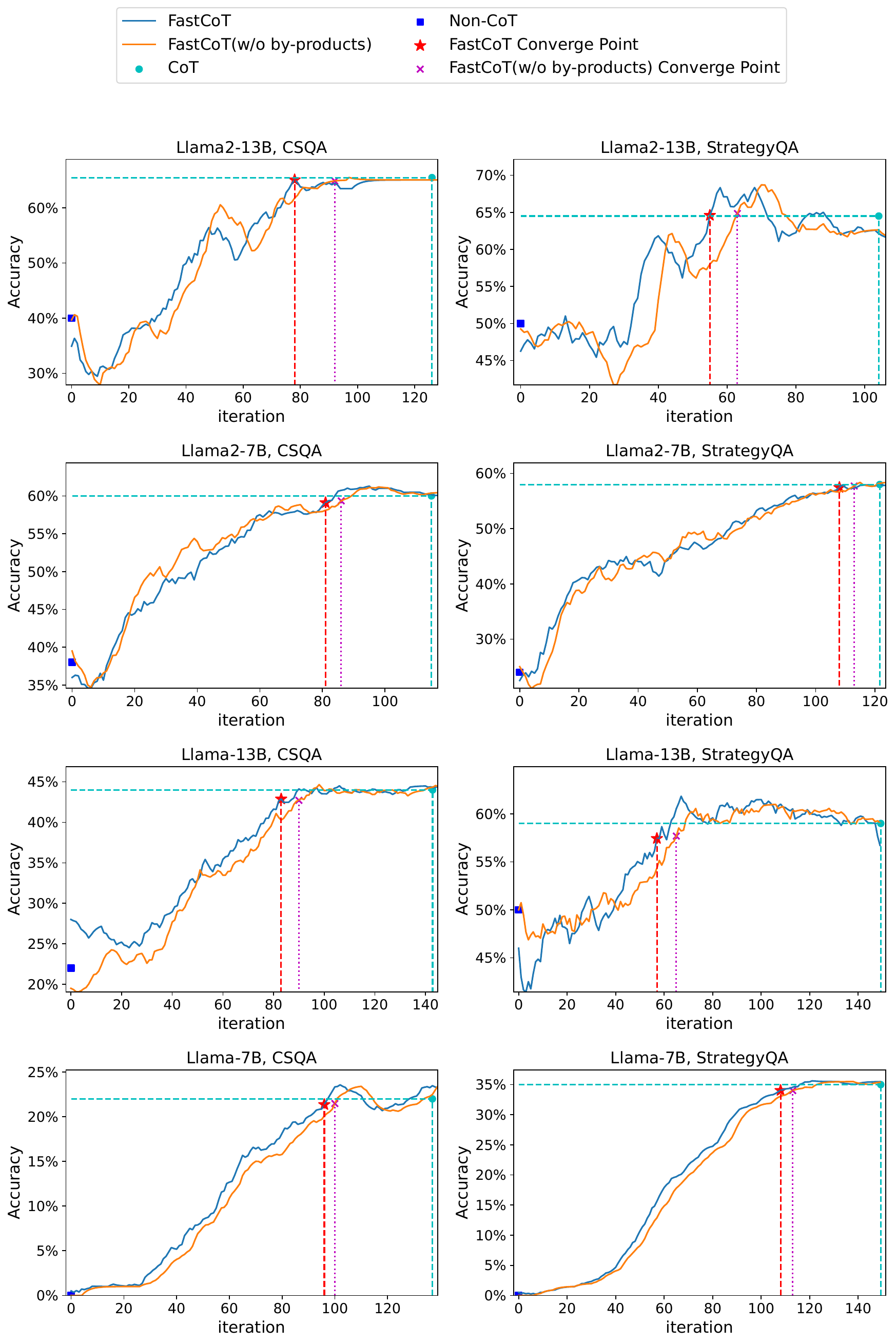}
    \caption{Multiple results of different models}
    \label{fig:multi_model results}
\end{figure}

First of all, to study in detail the performance of FastCoT with iterations, we truncate FastCoT(w/o by-products) and FastCoT in each iteration and conduct downstream CoT task experiments with an additional answer trigger. The main experiment results are shown in Figure \ref{fig:multi_model results}. Since approximate tokens are the only thing FastCoT has more than FastCoT(w/o by-products), a comparison between the two during the iterative process reveals that the roles played by approximate tokens in different tasks and different models are not entirely consistent. In terms of models, FastCoT almost always performs better than FastCoT (w/o by-products) on the CSQA data set and StrategyQA data set for Llama-13B, and Llama-7B. However, for the Llama2-13B and Llama2-7B models, the increasing trend of the curve is not perfect, and the performance of FastCoT and FastCoT (w/o by-products) has its advantages and disadvantages with iteration. But overall, FastCoT's performance is still better than FastCoT (w/o by-products).
It is worth noting that when the Llama2-13B model executes the StrategyQA question, whether it is FastCoT or FastCoT (w/o by-products), there is a phenomenon that the accuracy decreases as the number of precise rationale tokens increases within some iteration range. We believe that this phenomenon may also be related to the faithfulness of CoT\cite{lanham2023measuring, radhakrishnan2023question}.


\subsection{Time Analysis}

\subsubsection{Wall Clock time}

\begin{table*}[h]
\centering
{\small

\begin{tabular}{llcclllc}
\hline
Model                       & Dataset    & \multicolumn{1}{l}{FastCoT Time} & \multicolumn{1}{l}{CoT Time} & Save Time   &Time Ratio      & PL               & \multicolumn{1}{l}{IS/TI}               \\ \hline
\multirow{3}{*}{Llama2-13b} & CSQA       & 326.93s                          & 365.20s                      & 38.27s      & 10.47\%          & 1.27\%           & 896/14684                 \\
                            & AQUA       & 402.90s                          & 502.38s                      & 99.48s      & 19.80\%          & 2.66\%           & 1742/18640                \\
                            & StrategyQA & 313.43s                          & 330.38s                      & 16.95s      & 5.13\%          & 1.47\%           & 906/13992                 \\ \hline
\multirow{3}{*}{Llama2-7b}  & CSQA       & 184.98s                          & 242.47s                      & 57.49s      & 23.71\%          & 1.50\%           & 877/12866                 \\
                            & AQUA       & 228.04s                          & 257.46s                      & 29.42s      & 11.42\%          & 2.24\%           & 1400/15358                \\
                            & StrategyQA & 233.19s                          & 256.05s                      & 22.86s      & 8.92\%          & 2.21\%           & 1163/8958                 \\ \hline
\multirow{3}{*}{Llama1-13b} & CSQA       & 366.15s                          & 420.59s                      & 54.44s      & 13.00\%          & 1.84\%           & 986/16904                 \\
                            & AQUA       & 460.99s                          & 589.44s                      & 128.45s     & 21.80\%          & 2.50\%           & 1773/21674                \\
                            & StrategyQA & 293.38s                          & 372.71s                      & 79.33s      & 21.29\%          & 1.67\%           & 959/13036                 \\ \hline
\multirow{3}{*}{Llama1-7b}  & CSQA       & 286.30s                          & 297.13s                      & 10.83s      & 3.64\%          & 1.23\%           & 1216/18662                \\
                            & AQUA       & 348.87s                          & 388.48s                      & 39.64s      & 10.20\%          & 2.24\%           & 1807/23126                \\
                            & StrategyQA & 293.72s                          & 303.05s                      & 9.63s       & 3.20\%          & 2.01\%           & 1069/19216                \\ \hline
\end{tabular}

}
\caption{Wall clock time of the FastCoT. \textbf{PL} is the short for Performance Loss. }
\label{tab:time tabel}

\end{table*}

We conduct time overhead measurements on our method and the Vanilla CoT. The measurement results are shown in Table \ref{tab:time tabel}. In the IS/TI column of Table \ref{tab:time tabel}, we show the number of iterations saved by Jacobi decoding token hits and the total number of iterations. On most models and datasets, FastCoT achieves a significant improvement in terms of time varying from 3\% to 21.80\% at the cost of negligible performance drop within 3\%.


\subsubsection{Time Composition}

\begin{table}[h]
    \centering
    \begin{tabular}{lll}
        \hline
        Time Type      & FastCoT & AR       \\ \hline
        Inference     & 274.08s  & 358.15s   \\
        Type1 Padding         & 5.33s    & 0s        \\
        Type2 Padding         & 0.84s    & 0s        \\
        Decode         & 1.20s    & 3.06s     \\
        Context Decode & 5.42s    & 0s        \\
        Strip KV       & 18.58s   & 0s        \\
        Other          & 21.48s   & 3.99s     \\ 
        Total          & 326.93s  & 365.20s   \\ \hline
    \end{tabular}
    \caption{Time Composition}
    \label{tab:time composition}
\end{table}

\begin{figure}[h]
    \centering
    \includegraphics[width=1\linewidth]{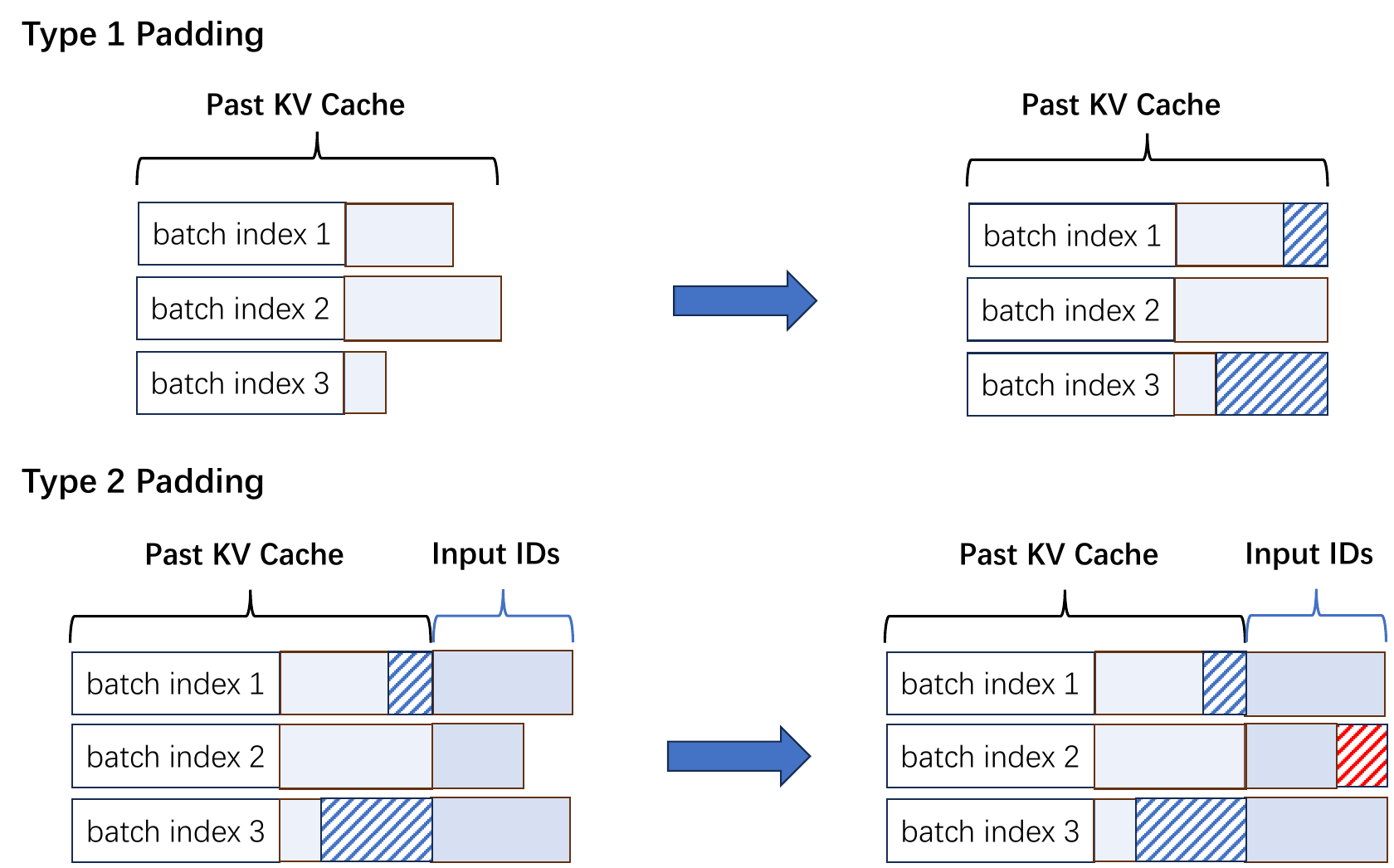}
    \caption{Two padding methods. Type1 Padding is employed to handle disparities in the lengths of historical KV caches. Type2 Padding is used to address inconsistencies in the number of tokens within Jacobi windows.}
    \label{fig: Padding}
\end{figure}

We perform further detailed analysis of the time cost of our method and compare it with the corresponding part of autoregressive decoding. The results are shown in Table \ref{tab:time composition}. We can see that if we only consider the time occupied by the GPU, which is the Inference time shown in the table. Since our method adds almost no additional overhead, but we can get the corresponding logits at the approximate tokens and the chance to save iteration cycles, the infer time was reduced by almost $30\%$.  However, calculating the approximate token in the context window after inference takes more 5.42s than AR decoding without this step. Moreover, due to the presence of Jacobi decoding, the positions of precise tokens within the same batch are almost always different, which results in varying storage lengths for the past key-value (KV) cache, necessitating additional padding to facilitate batch processing. To address this, we designed two types of padding mechanisms to cope with different scenarios as shown in Figure \ref{fig: Padding}. The respective time costs for Type1 and Type2 Padding are 5.33 seconds and 0.84 seconds, both negligible compared to the time spent on inference. Equally imperative within FastCoT is the handling of the key-value (KV) cache generated by the model's computation, and we call this procedure Strip KV, which consumes 18.58 seconds(about 5.6\%for total time) to extract the precise segment from the cache and retrieve it for subsequent iteration.


\subsection{Context Window Length}

\begin{figure}[h]
    \centering
    \includegraphics[width=1\linewidth]{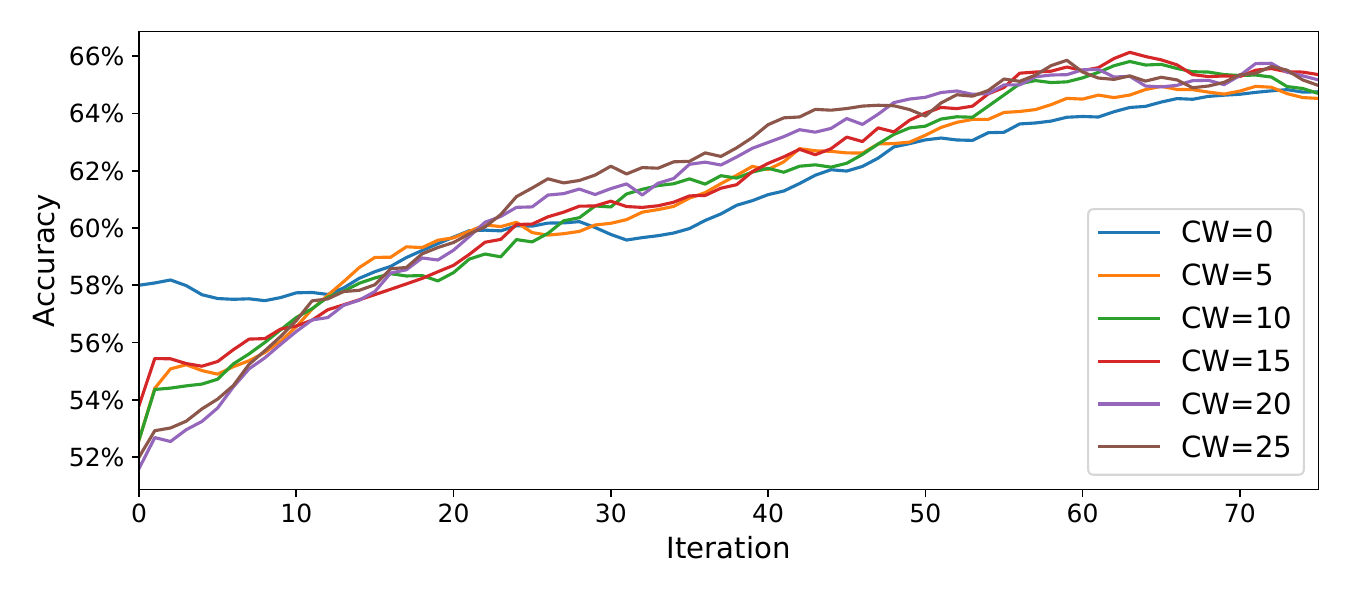}
    \caption{The performance of reasoning task with different context window through iterations}
    \label{fig: context window impact}
\end{figure}

In this section, we discuss the influence of the length of the Jacobi decoding Context Window on the performance of the reasoning task. As shown in Figure \ref{fig: context window impact}. We analyze the performance across multiple iteration stages. Initially, from iteration 0 to iteration 20, we observe that a longer context window correlates with poorer performance. A detailed analysis suggests that at the beginning of the iterations, the tokens decoded by Jacobi decoding may not be highly relevant to the question and can even have a negative impact. However, after iteration 10, with a sufficient number of iterations, the quality of the iterative solutions gradually improves, and the context window size begins to have a positive effect on the performance of the CoT task. During this phase, a larger context window is associated with better performance, but there is no significant difference observed for context windows larger than 20. By the time it reaches iteration 65, the performance gap between different context window sizes gradually diminishes.

In conclusion, the lower context window size works better than the higher one in the early stages of the iterations, we attribute this to the fact that the quality of the approximate tokens in the Jacobi context window decreases with the position in the context window. But after about 15 iterations, Jacobi decoding would generate enough informative tokens beneficial to downstream CoT tasks.

\section{Conclusion}
In conclusion, this paper introduces FastCoT, a model-agnostic framework that leverages Jacobi decoding to improve the efficiency of the CoT reasoning task. By providing the language model with approximate tokens as a glimpse of the future, FastCoT reduces the time overhead associated with autoregressive decoding. Through extensive experiments, it is demonstrated that FastCoT achieves a significant reduction in inference time, up to $20\%$, with only a negligible drop in performance compared to the regular approach. This study also presents one of the first attempts to introduce speedups in the CoT reasoning tasks, exploring accelerating the execution of reasoning class tasks from a generation perspective. We believe we pave the way for deeper future research.



\clearpage

\section{Limitations}
\label{sec:limitations}
Since our method involves modifying the text generation process by the model, it cannot be applied to black-box large language models. Furthermore, the choice of Context Window size is pre-defined and affected by the GPU and language model. Although we did not discuss it in our work, we believe that the process of controlling the context window throughout the iteration can be seen as a Markov Decision Process. It would be an intriguing problem to utilize reinforcement learning algorithms to regulate the context window size during the iteration while defining appropriate rewards. Another possible future work is how to accelerate the large language model's Jacobi decoding. Since Jacobi Decoding does not pursue completely consistent results with lossless decoding in most iterations, we can quickly conduct iterations by using a relatively smaller model for some iterations or Jacobi iterative tokens at positions to obtain better acceleration effects. Designing an algorithm in which the iteration of the Jacobi decoding part and the iteration of the lossless decoding part are not synchronized would also be a very interesting research direction.

\section{Ethical Considerations}
We utilized publicly available datasets to validate the accelerated application of our method in the CoT (Chain of Thought) scenarios. We adhered to the policies of the datasets used, without infringing on any copyright issues. And we believe that our research does not raise any additional ethical considerations.


\bibliography{anthology}

\clearpage











\end{document}